\documentclass{article}

\usepackage{arxiv}

\usepackage[utf8,latin9]{inputenc} 
\usepackage[T1]{fontenc}    
\usepackage{hyperref}       
\usepackage{url}            
\usepackage{booktabs}       
\usepackage{amsfonts}       
\usepackage{nicefrac}       
\usepackage{microtype}      
\usepackage{lipsum}

\usepackage{cite}
\usepackage{amsmath,amssymb,amsfonts}
\usepackage{algorithmic}
\usepackage{graphicx}
\usepackage{textcomp}
\usepackage{xcolor}
\usepackage{silence}
\usepackage{placeins}

\usepackage[T1]{fontenc}
\usepackage{multirow}
\PassOptionsToPackage{normalem}{ulem}
\usepackage{ulem}
\makeatletter

\providecommand{\tabularnewline}{\\}

\usepackage{graphicx}
\usepackage{colortbl}

\makeatother

\usepackage[linesnumbered,ruled,vlined]{algorithm2e}
\SetKwInput{kwInit}{Initialize} 
\usepackage{amsmath}
\DeclareMathOperator*{\argmax}{argmax} 
\usepackage{graphicx}

\title{WOTBoost: Weighted Oversampling Technique in Boosting for imbalanced learning}

\author{
  Wenhao Zhang\\
  University of California, Los Angeles\\
  \texttt{wenhaoz@ucla.edu} \\
   \And
  Ramin Ramezani \\
  University of California, Los Angeles \\
  \texttt{raminr@ucla.edu} \\
  \And
  Arash Naeim \\
  University of California, Los Angeles \\
  \texttt{ANaeim@mednet.ucla.edu} \\
}
\begin{document}
\maketitle

\begin{abstract}
Machine learning classifiers often stumble over imbalanced datasets where classes are not equally represented. This inherent bias towards the majority class may result in low accuracy in labeling minority class. Imbalanced learning is prevalent in many real-world applications, such as medical research, network intrusion detection, and fraud detection in credit card transactions, etc. A good number of research works have been reported to tackle this challenging problem. For example, Synthetic Minority Over-sampling TEchnique (SMOTE) and ADAptive SYNthetic sampling approach (ADASYN) use oversampling techniques to balance the skewed datasets. In this paper, we propose a novel method that combines a Weighted Oversampling Technique and ensemble Boosting method (WOTBoost) to improve the classification accuracy of minority data without sacrificing the accuracy of the majority class. WOTBoost adjusts its oversampling strategy at each round of boosting to synthesize more targeted minority data samples. The adjustment is enforced using a weighted distribution. We compare WOTBoost with other four classification models (i.e., decision tree, SMOTE + decision tree, ADASYN + decision tree, SMOTEBoost) extensively on 18 public accessible imbalanced datasets. WOTBoost achieves the best G mean on 6 datasets and highest AUC score on 7 datasets.
\end{abstract}


\section{Introduction}
Learning from imbalanced datasets can be very challenging as the classes are not equally represented in the datasets \cite{provost2000machine}. There might not be enough examples for a learner to form a legit hypothesis that can well model the under-represented classes. Hence, the classification results are often biased towards the majority classes. The curse of imbalanced learning is prevalent in real-world applications. In medical research, models are usually trained to give predictions on a dichotomous outcome based on a series of observable features \cite{elhassan2016classification}. For example, learning from a cancer dataset which mostly contains non-cancer data samples is perceived to be difficult. Other practical applications with more severely skewed datasets are fraudulent telephone calls \cite{fawcett1996combining}, detection of oil spills in satellite images \cite{kubat1998machine}, detection of network intrusions \cite{lee1998data}, and information retrieval and filtering tasks \cite{lewis1994heterogeneous}. In these scenarios, the imbalance ratio of majority class to minority class can go up to 100,000 \cite{chawla2002smote}. Even though class imbalance issue can exist in multi-class applications, we only focus on the binary class scenario in this paper as it is feasible to reduce a multi-class classification problem into a series of binary classification problems \cite{allwein2000reducing}.\\

To address the class imbalance issue, we proposed a novel method which combines a \textbf{W}eighted \textbf{O}versampling \textbf{T}echnique and ensemble \textbf{Boost}ing method (WOTBoost) to improve the classification accuracy of the minority data without sacrificing the accuracy of majority class. Essentially, the proposed method synthesizes data samples of minority class to balance the dataset. In addition, WOTBoost identifies the minority data samples which are mostly enclosed by the data samples from the other class. Empirically, it is deemed quite challenging to predict the true labels of these minority data samples. By placing enough synthesized data points within the proximity of the difficult minority data samples, the classification boundaries might be pushed away from the minority data samples. In other words, it is more likely to predict these difficult minorities to be minority class. Therefore,  WOTBoost creates more synthesized data points for these "difficult" minority data samples.

\noindent The contributions in this paper are as follows: 

\begin{itemize}
    \item  We identify the minority class data examples which are harder to learn at each round of boosting and generate more synthetic data for this kind.
    \item  We test our proposed algorithm extensively on 18 public accessible datasets and compared the results with the most commonly used algorithms. To our knowledge, this might be first work to carry out such a comprehensive comparison study in ensemble method combined with oversampling approach.
    \item We inspect the various distributions of 18 datasets and discussed why WOTBoost performs better on certain datasets.
\end{itemize}

\noindent The rest of the paper is organized as follows: section 2 briefly reviews the literature in dealing with imbalanced datasets. Section 3 proposes our WOTBoost algorithm with details. Section 4 compares the experimental results of WOTBoost algorithm with other baseline methods in terms of \textit{precision}, \textit{recall}, \textit{F measure},\textit{G mean}, \textit{Specificity}, \textit{AUC}. Section 5 discusses the results and propose the future work direction. 

\section{Background}


There have been ongoing efforts in this research domain finding ways to better tackle the imbalanced learning problem. Most of the state-of-the-art research methodologies are fallen into two major categories: 1) Data level approach, or 2) Algorithm level approach \cite{fernandez2018smote,ali2015classification}.

\subsection{Data level approach}
On the data level, skewed datasets can be balanced by either 1) oversampling the minority class data examples, 2) under-sampling the majority class data examples.

\subsubsection{Oversampling}
It aims to overcome the class imbalance by artificially creating new data from the under-represented class. However, simply duplicating the minority class samples would potentially cause overfitting. One of the most widely used techniques is SMOTE. The SMOTE algorithm generates synthetic data examples for minority class by randomly placing the newly created data instances between minority class data points and their neighbors \cite{chawla2002smote}. This technique not only can better model the minority classes by introducing a bias towards the minority instances but also has a lower chance of overfitting. This is due to SMOTE forcing the learners to create larger and less specific decision regions. Based on SMOTE, Hui Han et al. propose the Borderline-SMOTE, which only synthesizes the minorities on the decision borderline \cite{han2005borderline}. The Borderline-SMOTE classifies minority classes into "safe type" and "dangerous type". The "safe type" is located in the homogeneous regions where the majority of data examples belong to the same class. On the other hand, the "dangerous type" data points are outliers and most likely lie within the decision regions of the opposite class. The intention behind Borderline-SMOTE is to give more weights to the "dangerous type" minority class as it is deemed to be more difficult to learn \cite{napierala2016types}. Haibo He et al. adopt the same philosophy and proposed ADASYN algorithm, which uses a weighted distribution for different minority class data. The weights are assigned to minority data examples based on the level of difficulty in learning. In other words, harder data examples have more weights thus higher chance of getting more synthesized data. Prior to generating synthetic data, ADASYN inspects the $K$ nearest neighbors for each minority class data example, and counts the number of neighbors from the majority class, $\Delta_i$. Next, the difficulty of learning can be calculated as a ratio of $\Delta_i/K$ \cite{he2008adasyn}. ADASYN assigns higher weights on the difficult minority samples. On the contrary, Safe-Level-SMOTE gives more priority to safer minority instances and has a better accuracy performance than SMOTE and Borderline-SMOTE \cite{bunkhumpornpat2009safe}. Karia et al. propose a genetic algorithm, GenSample, for oversampling in imbalanced Datasets. GenSample accounts for the difficulty in learning minority examples when synthesizing, along with the performance improvement achieved by oversampling \cite{karia2019gensample}.

\subsubsection{Undersampling}
This technique approaches the imbalanced learning by removing a certain number of data examples from the majority class while keeping the original minority data points untouched. Random undersampling is the most common method in this category \cite{ali2015classification}. Elhassan AT et al. combine the undersampling algorithm with Tomek Link (T-Link) to create a balanced dataset \cite{elhassan2016classification,thai2010learning}. However, the undersampling method may suffer severe information loss. In this paper, we mainly focus on the oversampling technique and its variants \cite{tomek1976experiment}. 

\subsection{Algorithm level approach}
On the algorithm level, there are typically three mainstream approaches: a) Improved algorithms, b) cost-sensitive learning, and c) ensemble method \cite{ali2015classification,sun2009classification}. 

\subsubsection{Improved algorithms}
This approach generally attempts to tailor the classification algorithms to directly learn from the skewed dataset by shifting the decision boundary in favor of the minority class. Tasadduq Imam et al. propose z-SVM to counter the inherent bias in datasets by introducing a weight parameter, $z$, for minority class to correct the decision boundary during model fitting \cite{imam2006z}. Other modified SVM classifiers have also been reported, such as GSVM\_RU and BSVM \cite{tang2009svms,hoi2004biased}. One special form of an improved algorithm for imbalanced datasets is one-class learning. This method aims to generalize the hypothesis on a training dataset which only contains the target class \cite{manevitz2001one,devi2019learning}. 

\subsubsection{Cost-sensitive learning}
This technique penalizes the misclassifications of different classes with varying costs. Specifically, it assigns more costs to the misclassification of the target class. Hence, the false negative would be penalized more than the false positives \cite{zadrozny2001learning,margineantu2002class}. In cost-sensitive learning, a cost weight distribution is predefined in favor of the target classes. 

\subsubsection{Ensemble method}
Ensemble method trains a series of weak learners in a fixed number of iterations. A weak learner is a classifier whose accuracy is just barely above chance. At each round, a weak learner is created and a weak hypothesis is generalized. The predictive outcome is produced by aggregating all these weak hypotheses using a weighted voting method \cite{dietterich2000ensemble}. For example, AdaBoost.M2 algorithm calculates the pseudo-loss of each weak hypothesis during boosting. The pseudo-loss is computed over all data examples with respect to the incorrect classifications. The weight distribution is computed using the pseudo-loss (see algorithm 1). The weight distribution is updated with respect to pseudo loss at the current iteration and will be carried over to the next round of boosting. Hence, the learners in the next iteration will concentrate on the data examples which are hard to learn \cite{freund1996experiments}. Since Adaboost is apt to learn from a imbalanced dataset, several works are based on this boosting framework \cite{chawla2003smoteboost,seiffert2010rusboost,guo2004learning}. SMOTEBoost is proposed to combine the merits of SMOTE and Boosting methods by adding a SMOTE procedure at the beginning of each round of boosting. SMOTEBoost aims to improve the true positives without sacrificing the accuracy of majority class. RUSBoost alleviates class imbalanced by introducing random undersampling technique into a standard boosting procedure. Compared with SMOTEBoost, RUSBoost is a faster and simpler alternative to SMOTEBoost \cite{seiffert2010rusboost}. Ashutosh Kumar et al. proposed RUSTBoost algorithm which adds a redundancy-driven modified Tomek-Link based undersampling procedure before RUSBoost \cite{kumar2019improvement}. The Tomek-Link pairs are the pairs of closest data points from different classes. However, all the mentioned boosting algorithms treat the data examples equally. Krystyna Napierala et al. highlighted that the various types of minority data examples (e.g., safe, borderline, rare, and outlier) have unequal influence on the outcome of classification. As such, the algorithms should be designed to focus on the examples which are not easy to learn\cite{napierala2016types}. DataBoost-IM is reported to discriminate different types of data examples beforehand and adjust the weight distribution accordingly during boosting \cite{guo2004learning}.\\
\section{WOTBoost: Weighted Oversampling Technique in Boosting}

In this section, we propose the WOTBoost algorithm which combines a weighted oversampling algorithm with the standard boosting procedure. The Weighted Oversampling Technique populates synthetic data based on the weights that are associated to each minority data. In other words, higher weighted minority data samples are synthesized more. This algorithm is an ensemble method and creates a series of classifiers in an arbitrary number of iterations. The boosting procedure will be elaborated with details in Algorithm 1 and 2: a) We introduce a weighted oversampling step at the beginning of each iteration of boosting. b) We adjust the weighted oversampling strategy using the updated weights (i.e., $D_t$ at line 8 in algorithm 1) associated with the minority during each round of boosting \cite{provost2000machine}. The boosting algorithm gives more weights to the data samples which were misclassified in the previous round. Hence, WOTBoost can be designed to generate more synthetic data examples for the minority data which were misclassified in the previous iterations. Meanwhile, boosting technique would also add more weights to misclassified majority class data, and force the learner to focus on these data as well. Therefore, we combine the merits of weighted oversampling technique and AdaBoost.M2 together. The goal is to improve the discriminative power of the classifier on difficult minority examples without sacrificing the accuracy of the majority class data instances.\\

\begin{algorithm}[ht]
 \SetAlgoLined
 \KwIn{Training dataset $D_{Tr}$ with $m$ samples $\{x_i,y_i\}, i=1,2,...,m$, where $x_{i}$ is an instance in the $n$ dimensional feature space, $X$, and $y_i$ $\in Y=\{majority, minority\}$ is the label associated with $x_i$\; 
  Let $B=\{(i,\hat{y_i}): i=1,...,m, \hat{y_i} \neq y_i\}$\; $T$ specifies the number iterations in boosting procedure\;}
 \kwInit{$D_1(i, \hat{y_i})=\frac{1}{m}$, $i= 1,2,...,m$}
 \For{t=1,2,3,... $T$}{
  Create N synthetic examples from minority class with the weight distribution $D_t$ using \textbf{Algorithm \ref{algo:oversampling}}\;
  Fit a weak learner using the temporary training dataset which is a combination of original data and synthetic data\;
  Calculate a weak hypothesis $h_t: X \times Y \rightarrow [0,1]$\;
  Compute the pseudo-loss of $h_t$:
  $\varepsilon_t = \frac{1}{2}\sum\limits_{(i,\hat{y_i}) \in B}D_t(i,\hat{y_i})(1-h_t(x_i,y_i)+h_t(x_i,\hat{y_i}))$\;
  Let $\beta_t=\frac{\varepsilon_t}{1-\varepsilon_t}$\;
  Update the weight distribution $D_{t+1}$:
  $\tilde{D}_{t+1}(i,\hat{y_i})=D_t(i,\hat{y_i})\beta_t^{\frac{1}{2} \times (1-h_t(x_i,y_i)+h_t(x_i,\hat{y_i}))}$ \;
  Normalize $D_{t+1}:
  D_{t+1}(i,\hat{y_i}) = \frac{\tilde{D}_{t+1}(i,\hat{y_i})}{Z_t}$, where $Z_t$ is a normalization constant such that $\sum\limits_{i \in m}D_{t+1}(i,\hat{y_i})=1$}
 \KwOut{$h_{final}(x)=\argmax_{\hat{y_i} \in Y} \sum\limits_{t=1}^T(log\frac{1}{\beta_t})h_t(x,\hat{y_i})$
 }
 \caption{Boosting with weighted oversampling}
 \label{algo:boosting}
\end{algorithm}

\begin{algorithm}[h]
  \KwIn{N is the number of synthetic data examples from minority class\; $D_t$ is the weight distribution passed at \textit{line 2} in \textbf{Algorithm \ref{algo:boosting}}}
  Calculate the number of synthetic data examples for each minority class instance:
  $g_i = N \times \frac{D_t(i,\hat{y_i})}{\sum\limits_{j \in minority}{D_t(j, \hat{y_i})}}$ \;
  For each minority class instance, $x(i)$, in original training dataset, generate $g_i$ synthetic data examples using the following rules: \linebreak
  \For{ 1,2,3,...,$g_i$ }{
  Randomly choose a minority class example, $x_{nn}(i)$, from the $k$ nearest neighbors of $x(i)$, which is a n-dimensional feature vector \;
  Calculate the difference vector $\delta = x_{nn}(i) - x(i)$ \;
  Create a synthetic data example using the following equation
  \[x_{syn}(i) = x(i) + \delta \times \lambda\]
  where $\lambda \in [0,1]$
  }
  \KwOut{A temporary training dataset combining the original data with synthetic data}
  \caption{Dynamic weighted oversampling procedure}
  \label{algo:oversampling}
  
\end{algorithm}

\begin{figure*}[ht]
  \caption{Overview of the comparison study}
  \label{fig:flow}
  \centering
    \includegraphics[width=0.65\textwidth]{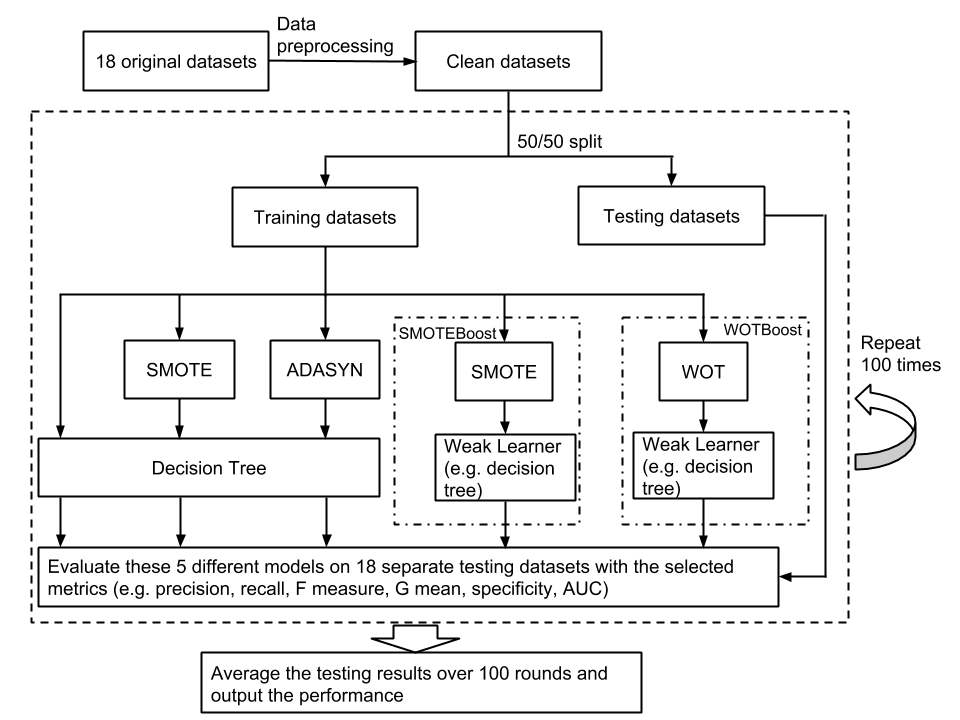}
\end{figure*}

\noindent Algorithm 1 presents the details of the boosting procedure, which is a modified version of AdaBoost.M2 \cite{freund1996experiments}. It takes a training dataset $D_{Tr}$ with $m$ data samples, $D_{Tr} = \{(x_1, y_1), (x_2,y_2),...,(x_m,y_m)\}$. $x_i$ is the \textit{ith} feature vector in $n$-dimensional space, and $y_i \in Y=\{majority, minority\}$ is the true label associated with $x_i$. $\hat{y_i}$ is the predicted label. We initialize a mislabel distribution, $B$, which contains all the misclassified data instances (i.e., $\hat{y_i} \neq y_i$).  In addition, we also initialize a weight distribution for the training data by assigning equal weights over all samples. During each round of boosting (step 1 - step 9), a weak learner is built on a training dataset which is the output of a weight oversampling procedure. The weak learner formulates a weak hypothesis which is just slightly better than random guessing, hence the name \cite{freund1996experiments}. But this is good enough as the final output will aggregate all the weak hypotheses using weighted voting. As for error estimation, the pseudo loss of a weak hypothesis is calculated as specified at step 5. Instead of using ordinary training loss, pseudo loss is adopted to force the ensemble method to focus on mislabeled data. More justification for using pseudo loss can be found in \cite{freund1996experiments,freund1997decision}. Once the pseudo loss is computed, the weight distribution, $D_t$, is updated accordingly and normalized at step 5 - step 8. \\

\noindent Algorithm 2 demonstrates the weighted oversampling procedure. The inputs to oversampling technique are the weight distribution, $D_{t}$, and an arbitrary number of synthetic data samples, $N$. It uses the weight distribution as the oversampling strategy to decide how to synthesize for each minority data samples, as it is described at step 1 in algorithm 2. As mentioned previously, the ensemble method would assign more weights to misclassified data. Therefore, this oversampling strategy facilitates the classifier to learn a broader representation of mislabeled data by placing more similar data samples around them.

\section{Experimentation}
In this section, we conduct a comprehensive comparison study of WOTBoost algorithm with decision tree, SMOTE + decision tree, ADASYN + decision tree, and SMOTEBoost. Figure \ref{fig:flow} shows how the models are built and assessed.

\begin{table*}[ht!]
\caption{Characteristics of 18 testing datasets}
\label{table:data}
\resizebox{1.0\textwidth}{!}{
  \centering
\begin{tabular}[c]{|l|c|c|c|c|c|c|c|}
\hline 
Dataset & Instances & Attributes & Outcome Frequency & Imbalanced Ratio & No. of safe minority & No. of unsafe minority & unsafe minority\%\tabularnewline
\hline 
Pima Indian Diabetes \cite{blake1998uci} & 768 & 9 & Maj: 506 Min:268 & 1.9 & 86 & 182 & 67.9\%\tabularnewline
\hline 
Abalone \cite{Dua:2019} & 4177 & 8 & Maj:689 Min:42 & 6.4 & 5 & 37 & 88.1\%\tabularnewline
\hline 
Vowel Recognition \cite{Dua:2019} & 990 & 14 & Maj:900 Min:90 & 10.0 & 89 & 1 & 1.1\%\tabularnewline
\hline 
Mammography \cite{elter2007prediction} & 11183 & 7 & Maj: 10923 Min: 260 & 42 & 107 & 153 & 58.8\%\tabularnewline
\hline 
Ionosphere \cite{Dua:2019} & 351 & 35 & Maj: 225 Min 126 & 1.8 & 57 & 69 & 54.8\%\tabularnewline
\hline 
Vehicle \cite{Dua:2019} & 846 & 19 & Maj: 647 Min:199 & 3.3 & 154 & 45 & 22.6\%\tabularnewline
\hline 
 Phoneme \cite{verleysen1995statistical} & 5404 & 6 & Ma j: 3818 Min:1580 & 2.4 & 980 & 606 & 38.2\%\tabularnewline
\hline 
Haberman \cite{Dua:2019} & 306 & 4 & Maj: 225 Min:81 & 2.8 & 8 & 73 & 90.1\%\tabularnewline
\hline 
Wisconsin \cite{Dua:2019} & 569 & 31 & Maj: 357 Min: 212 & 1.7 & 175 & 37 & 17.5\%\tabularnewline
\hline 
Blood Transfusion \cite{yeh2009knowledge} & 748 & 5 & Maj: 570 Min: 178 & 3.2 & 23 & 83 & 87.1\%\tabularnewline
\hline 
PC1 \cite{shirabad2005promise} & 1484 & 9 & Maj: 1032 Min: 77 & 13.4 & 8 & 69 & 89.6\%\tabularnewline
\hline 
Heart \cite{Dua:2019} & 294 & 14 & Maj: 188 Min: 106 & 1.8 & 17 & 89 & 84.0\%\tabularnewline
\hline 
Segment \cite{Dua:2019} & 2310 & 20 & Ma j: 1980 Min: 330 & 6.0 & 246 & 84 & 25.5\%\tabularnewline
\hline 
Yeast \cite{Dua:2019} & 1484 & 9 & Ma j: 1240 Min: 244 & 5.1 & 95 & 149 & 61.1\%\tabularnewline
\hline 
Oil & 937 & 50 & Maj: 896 Min: 41 & 21.9 & 0 & 41 & 100.0\%\tabularnewline
\hline 
Adult  \cite{Dua:2019} & 48842 & 7 & Maj: 37155 Min: 11687 & 3.2 & 873 & 10814 & 92.5\%\tabularnewline
\hline 
Satimage \cite{Dua:2019} & 6430 & 37 & Maj: 5805 Min: 625 & 9.3 & 328 & 297 & 47.5\%\tabularnewline
\hline 
Forest cover \cite{blackard1999comparative} & 581012 & 11 & Maj: 35754 Min: 2747 & 13.0 & 2079 & 668 & 24.3\%\tabularnewline
\hline 
\end{tabular}
}
\end{table*}

\label{section:datasets}

\subsection{Dataset overview}
We evaluate these 5 models extensively using 18 imbalanced datasets which are publicly accessible. The imbalanced ratio (i.e., counts of majority class samples to counts of minority class samples) of these datasets vary from 1.7 to 42. Since some testing imbalanced datasets have more than 2 classes, and we are only interested in the binary class problem in this paper, we pre-processed these datasets and modified them into a binary class datasets following the rules in the literature \cite{chawla2002smote,han2005borderline,bunkhumpornpat2009safe,he2008adasyn,chawla2003smoteboost,kumar2019improvement}. Meanwhile, only numeric attributes are included when processing datasets. The details of data cleaning can be referred to the prior works \cite{chawla2002smote, chawla2003smoteboost, he2008adasyn}. The characteristics of these datasets are summarized in table \ref{table:data}.

\subsection{Experiment setup}
We compare the WOTBoost algorithm with naive decision tree classifier, decision tree classifier after SMOTE, decision tree classifier after ADASYN, and SMOTEBoost. Figure \ref{fig:flow} shows that the clean datasets are split evenly into training and testing during each iteration \cite{he2008adasyn}. As a control group, a naive decision tree model learned directly from the imbalanced training dataset. SMOTE and ADASYN algorithms are used separately to balance the training dataset before inputting it to decision tree classifiers. SMOTEBoost and WOTBoost take in imbalanced training datasets and synthesize new data samples for the minority at each round of boosting. Both of them use decision tree as the weak learner \cite{chawla2003smoteboost}. Models are evaluated on a separate testing dataset. The evaluating metrics used in this study are precision, recall, F1 measure, G mean, specificity, area under ROC. The final performance assessments are averaged over 100 such runs, and they are summarized in table \ref{table:performance}. During each testing run, we oversample the training dataset in a way that both minority class and majority class are equally represented in all models \cite{he2008adasyn}. For SMOTE, ADASYN, SMOTEBoost, and WOTBoost, we set the number of nearest neighbors to be 5.\\

\subsection{Metrics}
\noindent Overall accuracy is typically chosen to evaluate the predictive power of machine learning classifiers provided with a balanced dataset. As for imbalanced datasets, overall accuracy is no longer an effective metric. For example, in the information retrieval and filtering domain by Lewis and Catlette (1994), only 0.2\% are interesting cases \cite{kubat1998machine}. A dummy classifier that always gives predictions of majority class would easily achieve an overall accuracy of 99.8\%. However, this predictive model is uninformative as we are more interested in classifying the minority class. Common alternatives to overall accuracy in assessing imbalanced learning models are F measures, G mean, and Area Under the Curve (AUC) for Receiver Operating Characteristic (ROC) \cite{swets1988measuring}. By convention, majority class is regarded as negative class and minority class as positive class \cite{chawla2002smote,kubat1997addressing}. Table II shows a confusion matrix that is typically used to visualize and assess the performance of predictive models. Based on this confusion matrix, the evaluation metrics used in this paper are mathematically formulated as follows:


\begin{table}[]
\label{table:confusion}
\caption{Confusion matrix of a binary classification problem}
\centering
\begin{tabular}{|l|l|l|}
\hline
                                                                      & \textbf{\begin{tabular}[c]{@{}l@{}}Actual Positive\end{tabular}} & \textbf{\begin{tabular}[c]{@{}l@{}}Actual Negative\end{tabular}} \\ \hline
\textbf{\begin{tabular}[c]{@{}l@{}}Predicted Positive\end{tabular}} & True Positive (TP)                                                 & False Positive (FP)                                                \\ \hline
\textbf{\begin{tabular}[c]{@{}l@{}}Predicted Negative\end{tabular}} & False Negative (FN)                                                & True Negative (TN)                                                 \\ \hline
\end{tabular}
\end{table}

\begin{equation}
\small  Precision = \frac{TP}{TP+FP}
\end{equation}

\begin{equation}
\small  Recall = \frac{TP}{TP+FN}
\end{equation}

\begin{equation}
\small  F1\ measure = 2 \cdot \frac{precision \times recall}{precision + recall}
\end{equation}

\begin{equation}
\small
\label{eq:gmean}
  \begin{split}
  G\ mean &= \sqrt{Positive\ Accuracy \times Negative\ Accuracy}\\
  &= \sqrt{\frac{TP}{TP+FN} \times \frac{TN}{TN+FP}}
\end{split}
\end{equation}

\begin{equation}
\small Specificity = \frac{TN}{TN+FP}
\end{equation}

\section{Results}

\begin{table*}[!ht]
\caption{Evaluation metrics and performance comparison}

\noindent\resizebox{\textwidth}{!}{

\begin{tabular}{|c|c|c|c|c|c|c|c|c|c|c|c|}
\hline 
Dataset & Methods$^a$ & OA$^b$ & Precision$^b$ & Recall$^b$ & F\_measure$^b$ & G\_mean$^b$ & Specificity$^b$ & Sensitivity$^b$ & ROC AUC$^b$ & Outcome Frequency & Imbalanced ratio\tabularnewline
\hline 
\hline 
\multirow{5}{*}{Pima Indian Diabetes} & DT & 0.71 $\pm$ 0.02 & \textbf{\uline{0.61 $\pm$ 0.04}} & 0.54 $\pm$ 0.05& 0.57 $\pm$ 0.03& 0.66 $\pm$ 0.02& \textbf{\uline{0.80 $\pm$ 0.03}} & 0.54 $\pm$ 0.05& 0.67$\pm$0.02 & \multirow{5}{*}{Maj: 500 Min:268} & \multirow{5}{*}{1.9}\tabularnewline
\cline{2-10} \cline{3-10} \cline{4-10} \cline{5-10} \cline{6-10} \cline{7-10} \cline{8-10} \cline{9-10} \cline{10-10} 
 & S & 0.67$\pm$0.02 & 0.55$\pm$0.03 & 0.54$\pm$0.04 & 0.54$\pm$0.02 & 0.63$\pm$0.02 & 0.75$\pm$0.03 & 0.54$\pm$0.04 & 0.64$\pm$0.02 &  & \tabularnewline
\cline{2-10} \cline{3-10} \cline{4-10} \cline{5-10} \cline{6-10} \cline{7-10} \cline{8-10} \cline{9-10} \cline{10-10} 
 & A & 0.68 $\pm$ 0.02 & 0.56$\pm$0.04 & 0.58$\pm$0.05 & 0.57$\pm$0.03 & 0.66$\pm$0.03 & 0.74$\pm$0.03 & 0.58$\pm$0.05 & 0.66$\pm$0.02 &  & \tabularnewline
\cline{2-10} \cline{3-10} \cline{4-10} \cline{5-10} \cline{6-10} \cline{7-10} \cline{8-10} \cline{9-10} \cline{10-10} 
 & SM & 0.66$\pm$0.02 & 0.52$\pm$0.02 & \textbf{\uline{0.86$\pm$0.04}} & 0.64$\pm$0.02 & 0.68$\pm$0.02 & 0.54$\pm$0.05 & \textbf{\uline{0.86$\pm$0.04}} & 0.70$\pm$0.01 &  & \tabularnewline
\cline{2-10} \cline{3-10} \cline{4-10} \cline{5-10} \cline{6-10} \cline{7-10} \cline{8-10} \cline{9-10} \cline{10-10} 
 & WOT & \textbf{\uline{0.73$\pm$0.02}} & 0.60$\pm$0.03 & 0.78$\pm$0.05 & \textbf{\uline{0.68$\pm$0.02}} & \textbf{\uline{0.74$\pm$0.02}} & 0.71$\pm$0.03 & 0.78$\pm$0.05 & \textbf{\uline{0.74$\pm$0.02}} &  & \tabularnewline
\hline 
\multirow{5}{*}{Abalone} & DT & 0.93 $\pm$ 0.01 & 0.46$\pm$0.12 & \textbf{\uline{0.46$\pm$0.10}} & \textbf{\uline{0.46$\pm$0.08}} & \textbf{\uline{0.66$\pm$0.08}} & 0.96$\pm$0.01 & \textbf{\uline{0.46$\pm$0.10}} & \textbf{\uline{0.71$\pm$0.04}} & \multirow{5}{*}{Maj:689 Min:42} & \multirow{5}{*}{16.4}\tabularnewline
\cline{2-10} \cline{3-10} \cline{4-10} \cline{5-10} \cline{6-10} \cline{7-10} \cline{8-10} \cline{9-10} \cline{10-10} 
 & S & 0.88$\pm$0.02 & 0.24$\pm$0.07 & 0.38$\pm$0.11 & 0.29$\pm$0.07 & 0.59$\pm$0.08 & 0.92$\pm$0.02 & 0.38$\pm$0.11 & 0.65$\pm$0.05 &  & \tabularnewline
\cline{2-10} \cline{3-10} \cline{4-10} \cline{5-10} \cline{6-10} \cline{7-10} \cline{8-10} \cline{9-10} \cline{10-10} 
 & A & 0.88$\pm$0.02 & 0.24$\pm$0.06 & 0.42$\pm$0.11 & 0.31$\pm$0.07 & 0.62$\pm$0.09 & 0.91$\pm$0.02 & 0.42$\pm$0.11 & 0.66$\pm$0.05 &  & \tabularnewline
\cline{2-10} \cline{3-10} \cline{4-10} \cline{5-10} \cline{6-10} \cline{7-10} \cline{8-10} \cline{9-10} \cline{10-10} 
 & SM & 0.84$\pm$0.06 & 0.19$\pm$0.04 & \textbf{\uline{0.46$\pm$0.12}} & 0.27$\pm$0.05 & 0.63$\pm$0.05 & 0.87$\pm$0.07 & \textbf{\uline{0.46$\pm$0.12}} & 0.66$\pm$0.05 &  & \tabularnewline
\cline{2-10} \cline{3-10} \cline{4-10} \cline{5-10} \cline{6-10} \cline{7-10} \cline{8-10} \cline{9-10} \cline{10-10} 
 & WOT & \textbf{\uline{0.94$\pm$0.01}} & \textbf{\uline{0.55$\pm$0.33}} & 0.34$\pm$0.11 & 0.42$\pm$0.13 & 0.58$\pm$ 0.18& \textbf{\uline{0.98$\pm$0.01}} & 0.34 $\pm$0.11& 0.66 $\pm$0.05&  & \tabularnewline
\hline 
\multirow{5}{*}{Vowel Recognition} & DT & 0.97$\pm$0.00 & \textbf{\uline{0.90$\pm$0.06}} & 0.79$\pm$0.06 & 0.84$\pm$0.04 & 0.88$\pm$0.03 & \textbf{\uline{0.99$\pm$0.00}} & 0.79$\pm$0.06 & 0.89$\pm$0.03 & \multirow{5}{*}{Maj:900 Min:90} & \multirow{5}{*}{10.0}\tabularnewline
\cline{2-10} \cline{3-10} \cline{4-10} \cline{5-10} \cline{6-10} \cline{7-10} \cline{8-10} \cline{9-10} \cline{10-10} 
 & S & 0.96$\pm$0.00 & 0.85$\pm$0.06 & 0.74$\pm$0.06 & 0.80$\pm$0.04 & 0.86$\pm$0.03 & 0.99$\pm$0.00 & 0.74$\pm$0.06 & 0.87$\pm$0.03 &  & \tabularnewline
\cline{2-10} \cline{3-10} \cline{4-10} \cline{5-10} \cline{6-10} \cline{7-10} \cline{8-10} \cline{9-10} \cline{10-10} 
 & A & 0.97$\pm$0.00 & 0.88$\pm$0.05 & 0.79$\pm$0.07 & 0.83$\pm$0.04 & 0.88$\pm$0.03 & 0.99$\pm$0.00 & 0.79$\pm$0.07 & 0.89$\pm$0.03 &  & \tabularnewline
\cline{2-10} \cline{3-10} \cline{4-10} \cline{5-10} \cline{6-10} \cline{7-10} \cline{8-10} \cline{9-10} \cline{10-10} 
 & SM & \textbf{\uline{0.98$\pm$0.00}} & 0.83$\pm$0.05 & 0.96$\pm$0.04 & 0.89$\pm$0.03 & 0.97$\pm$0.02 & 0.98$\pm$0.00 & 0.96$\pm$0.04 & 0.97$\pm$0.02 &  & \tabularnewline
\cline{2-10} \cline{3-10} \cline{4-10} \cline{5-10} \cline{6-10} \cline{7-10} \cline{8-10} \cline{9-10} \cline{10-10} 
 & WOT & \textbf{\uline{0.98$\pm$0.01}} & 0.87$\pm$0.10 & \textbf{\uline{0.98$\pm$0.01}} & \textbf{\uline{0.93$\pm$0.07}} & \textbf{\uline{0.98$\pm$0.02}} & \textbf{\uline{0.99$\pm$0.01}} & \textbf{\uline{0.98$\pm$0.01}} & \textbf{\uline{0.98$\pm$0.01}} &  & \tabularnewline
\hline 
\multirow{5}{*}{Ionosphere} & DT & 0.86$\pm$0.02 & 0.83$\pm$0.06 & 0.73$\pm$0.06 & 0.77$\pm$0.04 & 0.82$\pm$0.03 & 0.92$\pm$0.04 & 0.73$\pm$0.06 & 0.83$\pm$0.03 & \multirow{5}{*}{Maj: 225 Min 126} & \multirow{5}{*}{1.8}\tabularnewline
\cline{2-10} \cline{3-10} \cline{4-10} \cline{5-10} \cline{6-10} \cline{7-10} \cline{8-10} \cline{9-10} \cline{10-10} 
 & S & 0.85$\pm$0.03 & 0.75$\pm$0.05 & 0.81$\pm$0.06 & 0.78$\pm$0.04 & 0.84$\pm$0.03 & 0.86$\pm$0.03 & 0.81$\pm$0.06 & 0.84$\pm$0.03 &  & \tabularnewline
\cline{2-10} \cline{3-10} \cline{4-10} \cline{5-10} \cline{6-10} \cline{7-10} \cline{8-10} \cline{9-10} \cline{10-10} 
 & A & 0.88$\pm$0.03 & 0.84$\pm$0.05 & 0.80$\pm$0.06 & 0.82$\pm$0.04 & 0.86$\pm$0.03 & 0.92$\pm$0.03 & 0.80$\pm$0.06 & 0.86$\pm$0.03 &  & \tabularnewline
\cline{2-10} \cline{3-10} \cline{4-10} \cline{5-10} \cline{6-10} \cline{7-10} \cline{8-10} \cline{9-10} \cline{10-10} 
 & SM & \textbf{\uline{0.91$\pm$0.02}} & 0.89$\pm$0.06 & \textbf{\uline{0.85$\pm$0.04}} & \textbf{\uline{0.87$\pm$0.03}} & \textbf{\uline{0.90$\pm$0.02}} & 0.95$\pm$0.04 & \textbf{\uline{0.85$\pm$0.04}} & \textbf{\uline{0.90$\pm$0.02}} &  & \tabularnewline
\cline{2-10} \cline{3-10} \cline{4-10} \cline{5-10} \cline{6-10} \cline{7-10} \cline{8-10} \cline{9-10} \cline{10-10} 
 & WOT & \textbf{\uline{0.91$\pm$0.02}} & \textbf{\uline{0.92$\pm$0.05}} & 0.79$\pm$0.04 & 0.85$\pm$0.03 & 0.87$\pm$0.02 & \textbf{\uline{0.97$\pm$0.02}} & 0.79$\pm$0.04 & 0.88$\pm$0.02 &  & \tabularnewline
\hline 
\multirow{5}{*}{Vehicle} & DT & 0.94$\pm$0.01 & \textbf{\uline{0.85$\pm$0.04}} & 0.88$\pm$0.04 & 0.87$\pm$0.03 & 0.92$\pm$0.02 & \textbf{\uline{0.95$\pm$0.01}} & 0.88$\pm$0.04 & 0.92$\pm$0.02 & \multirow{5}{*}{Maj: 647 Min:199} & \multirow{5}{*}{3.3}\tabularnewline
\cline{2-10} \cline{3-10} \cline{4-10} \cline{5-10} \cline{6-10} \cline{7-10} \cline{8-10} \cline{9-10} \cline{10-10} 
 & S & 0.90$\pm$0.01 & 0.75$\pm$0.04 & 0.88$\pm$0.05 & 0.81$\pm$0.03 & 0.89$\pm$0.02 & 0.91$\pm$0.01 & 0.88$\pm$0.05 & 0.89$\pm$0.02 &  & \tabularnewline
\cline{2-10} \cline{3-10} \cline{4-10} \cline{5-10} \cline{6-10} \cline{7-10} \cline{8-10} \cline{9-10} \cline{10-10} 
 & A & 0.92$\pm$0.01 & 0.81$\pm$0.04 & 0.87$\pm$0.04 & 0.84$\pm$0.02 & 0.90$\pm$0.02 & 0.93$\pm$0.01 & 0.87$\pm$0.04 & 0.90$\pm$0.02 &  & \tabularnewline
\cline{2-10} \cline{3-10} \cline{4-10} \cline{5-10} \cline{6-10} \cline{7-10} \cline{8-10} \cline{9-10} \cline{10-10} 
 & SM & \textbf{\uline{0.95$\pm$0.00}} & 0.84$\pm$0.03 & \textbf{\uline{0.97$\pm$0.02}} & \textbf{\uline{0.90$\pm$0.02}} & \textbf{\uline{0.96$\pm$0.01}} & 0.94$\pm$0.01 & \textbf{\uline{0.97$\pm$0.02}} & \textbf{\uline{0.96$\pm$0.01}} &  & \tabularnewline
\cline{2-10} \cline{3-10} \cline{4-10} \cline{5-10} \cline{6-10} \cline{7-10} \cline{8-10} \cline{9-10} \cline{10-10} 
 & WOT & 0.89$\pm$0.10 & 0.70$\pm$0.15 & \textbf{\uline{0.97$\pm$0.03}} & 0.81$\pm$0.11 & 0.92$\pm$0.07 & 0.87$\pm$0.14 & \textbf{\uline{0.97$\pm$0.03}} & 0.92$\pm$0.06 &  & \tabularnewline
\hline 
\multirow{5}{*}{Phoneme} & DT & \textbf{\uline{0.86$\pm$0.00}} & \textbf{\uline{0.75$\pm$0.01}} & 0.74$\pm$0.01 & 0.75$\pm$0.01 & 0.82$\pm$0.00 & \textbf{\uline{0.90$\pm$0.00}} & 0.74$\pm$0.01 & 0.82$\pm$0.00 & \multirow{5}{*}{Maj: 3818 Min:1580} & \multirow{5}{*}{2.4}\tabularnewline
\cline{2-10} \cline{3-10} \cline{4-10} \cline{5-10} \cline{6-10} \cline{7-10} \cline{8-10} \cline{9-10} \cline{10-10} 
 & S & \textbf{\uline{0.86$\pm$0.00}} & 0.74$\pm$0.01 & 0.78$\pm$0.01 & \textbf{\uline{0.76$\pm$0.01}} & \textbf{\uline{0.83$\pm$0.01}} & 0.89$\pm$0.00 & 0.78$\pm$0.01 & \textbf{\uline{0.83$\pm$0.00}} &  & \tabularnewline
\cline{2-10} \cline{3-10} \cline{4-10} \cline{5-10} \cline{6-10} \cline{7-10} \cline{8-10} \cline{9-10} \cline{10-10} 
 & A & 0.83$\pm$0.00 & 0.68$\pm$0.01 & 0.78$\pm$0.01 & 0.73$\pm$0.01 & 0.82$\pm$0.00 & 0.85$\pm$0.00 & 0.78$\pm$0.01 & 0.82$\pm$0.00 &  & \tabularnewline
\cline{2-10} \cline{3-10} \cline{4-10} \cline{5-10} \cline{6-10} \cline{7-10} \cline{8-10} \cline{9-10} \cline{10-10} 
 & SM & 0.77$\pm$0.00 & 0.57$\pm$0.01 & 0.86$\pm$0.01 & 0.69$\pm$0.01 & 0.80$\pm$0.00 & 0.74$\pm$0.01 & 0.86$\pm$0.01 & 0.80$\pm$0.00 &  & \tabularnewline
\cline{2-10} \cline{3-10} \cline{4-10} \cline{5-10} \cline{6-10} \cline{7-10} \cline{8-10} \cline{9-10} \cline{10-10} 
 & WOT & 0.52$\pm$0.06 & 0.38$\pm$0.03 & \textbf{\uline{0.99$\pm$0.01}} & 0.54$\pm$0.03 & 0.57$\pm$0.07 & 0.34$\pm$0.09 & \textbf{\uline{0.99$\pm$0.01}} & 0.66$\pm$0.04 &  & \tabularnewline
\hline 
\multirow{5}{*}{Haberman} & DT & \textbf{\uline{0.67$\pm$0.03}} & 0.38$\pm$0.06 & 0.25$\pm$0.08 & 0.30$\pm$0.05 & 0.46$\pm$0.05 & \textbf{\uline{0.83$\pm$0.05}} & 0.25$\pm$0.08 & 0.54$\pm$0.03 & \multirow{5}{*}{Maj: 225 Min:81} & \multirow{5}{*}{2.8}\tabularnewline
\cline{2-10} \cline{3-10} \cline{4-10} \cline{5-10} \cline{6-10} \cline{7-10} \cline{8-10} \cline{9-10} \cline{10-10} 
 & S & 0.65$\pm$0.03 & \textbf{\uline{0.40$\pm$0.05}} & 0.39$\pm$0.08 & 0.39$\pm$0.05 & \textbf{\uline{0.64$\pm$0.04}} & 0.76$\pm$0.05 & 0.39$\pm$0.08 & 0.57$\pm$0.03 &  & \tabularnewline
\cline{2-10} \cline{3-10} \cline{4-10} \cline{5-10} \cline{6-10} \cline{7-10} \cline{8-10} \cline{9-10} \cline{10-10} 
 & A & 0.60$\pm$0.03 & 0.37$\pm$0.05 & 0.52$\pm$0.08 & 0.43$\pm$0.05 & 0.58$\pm$0.05 & 0.63$\pm$0.05 & 0.52$\pm$0.08 & 0.58$\pm$0.04 &  & \tabularnewline
\cline{2-10} \cline{3-10} \cline{4-10} \cline{5-10} \cline{6-10} \cline{7-10} \cline{8-10} \cline{9-10} \cline{10-10} 
 & SM & 0.48$\pm$0.06 & 0.34$\pm$0.03 & \textbf{\uline{0.84$\pm$0.07}} & \textbf{\uline{0.48$\pm$0.03}} & 0.53$\pm$0.10 & 0.34$\pm$0.11 & \textbf{\uline{0.84$\pm$0.07}} & \textbf{\uline{0.59$\pm$0.02}} &  & \tabularnewline
\cline{2-10} \cline{3-10} \cline{4-10} \cline{5-10} \cline{6-10} \cline{7-10} \cline{8-10} \cline{9-10} \cline{10-10} 
 & WOT & 0.54$\pm$0.05 & 0.35$\pm$0.07 & 0.70$\pm$0.12 & 0.47$\pm$0.05 & 0.57$\pm$0.05 & 0.48$\pm$0.11 & 0.70$\pm$0.12 & \textbf{\uline{0.59$\pm$0.03}} &  & \tabularnewline
\hline 
\multirow{5}{*}{Wisconsin} & DT & 0.95$\pm$0.01 & 0.93$\pm$0.03 & 0.93$\pm$0.01 & 0.93$\pm$0.01 & 0.95$\pm$0.01 & 0.96$\pm$0.02 & 0.93$\pm$0.01 & 0.95$\pm$0.01 & \multirow{5}{*}{Maj: 357 Min: 212} & \multirow{5}{*}{1.7}\tabularnewline
\cline{2-10} \cline{3-10} \cline{4-10} \cline{5-10} \cline{6-10} \cline{7-10} \cline{8-10} \cline{9-10} \cline{10-10} 
 & S & 0.92$\pm$0.01 & 0.89$\pm$0.03 & 0.90 $\pm$0.03& 0.89$\pm$0.02 & 0.91$\pm$0.01 & 0.93$\pm$0.02 & 0.90$\pm$0.03 & 0.91$\pm$0.01 &  & \tabularnewline
\cline{2-10} \cline{3-10} \cline{4-10} \cline{5-10} \cline{6-10} \cline{7-10} \cline{8-10} \cline{9-10} \cline{10-10} 
 & A & 0.95$\pm$0.01 & 0.93$\pm$0.03 & 0.94$\pm$0.03 & 0.94$\pm$0.02 & 0.95$\pm$0.01 & 0.96$\pm$0.02 & 0.94$\pm$0.03 & 0.95$\pm$0.01 &  & \tabularnewline
\cline{2-10} \cline{3-10} \cline{4-10} \cline{5-10} \cline{6-10} \cline{7-10} \cline{8-10} \cline{9-10} \cline{10-10} 
 & SM & \textbf{\uline{0.98$\pm$0.01}} & \textbf{\uline{0.99$\pm$0.00}} & \textbf{\uline{0.95$\pm$0.01}} & \textbf{\uline{0.97$\pm$0.01}} & \textbf{\uline{0.97$\pm$0.01}} & \textbf{\uline{0.99$\pm$0.00}} & \textbf{\uline{0.95$\pm$0.01}} & \textbf{\uline{0.97$\pm$0.01}} &  & \tabularnewline
\cline{2-10} \cline{3-10} \cline{4-10} \cline{5-10} \cline{6-10} \cline{7-10} \cline{8-10} \cline{9-10} \cline{10-10} 
 & WOT & 0.97$\pm$0.01 & 0.97$\pm$0.03 & \textbf{\uline{0.95$\pm$0.02}} & 0.96$\pm$0.02 & 0.96$\pm$0.01 & 0.98$\pm$0.02 & \textbf{\uline{0.95$\pm$0.02}} & 0.96$\pm$0.01 &  & \tabularnewline
\hline 
\multirow{5}{*}{Blood Transfusion} & DT & \textbf{\uline{0.72$\pm$0.01}} & \textbf{\uline{0.39$\pm$0.06}} & 0.28$\pm$0.08 & 0.32$\pm$0.06 & 0.49$\pm$0.07 & \textbf{\uline{0.86$\pm$0.01}} & 0.28$\pm$0.08 & 0.57$\pm$0.04 & \multirow{5}{*}{Maj: 570 Min: 178} & \multirow{5}{*}{3.2}\tabularnewline
\cline{2-10} \cline{3-10} \cline{4-10} \cline{5-10} \cline{6-10} \cline{7-10} \cline{8-10} \cline{9-10} \cline{10-10} 
 & S & 0.71$\pm$0.01 & \textbf{\uline{0.39$\pm$0.05}} & 0.39$\pm$0.07 & 0.39$\pm$0.05 & 0.56$\pm$0.05 & 0.81$\pm$0.01 & 0.39$\pm$0.07 & 0.60$\pm$0.03 &  & \tabularnewline
\cline{2-10} \cline{3-10} \cline{4-10} \cline{5-10} \cline{6-10} \cline{7-10} \cline{8-10} \cline{9-10} \cline{10-10} 
 & A & 0.70$\pm$0.01 & 0.38$\pm$0.05 & 0.42$\pm$0.08 & 0.40$\pm$0.06 & 0.57$\pm$0.07 & 0.78$\pm$0.01 & 0.42$\pm$0.08 & 0.60$\pm$0.04 &  & \tabularnewline
\cline{2-10} \cline{3-10} \cline{4-10} \cline{5-10} \cline{6-10} \cline{7-10} \cline{8-10} \cline{9-10} \cline{10-10} 
 & SM & 0.44$\pm$0.03 & 0.29$\pm$0.03 & \textbf{\uline{0.93$\pm$0.10}} & \textbf{\uline{0.45$\pm$0.03}} & 0.52$\pm$0.04 & 0.29$\pm$0.04 & \textbf{\uline{0.93$\pm$0.10}} & 0.61$\pm$0.04 &  & \tabularnewline
\cline{2-10} \cline{3-10} \cline{4-10} \cline{5-10} \cline{6-10} \cline{7-10} \cline{8-10} \cline{9-10} \cline{10-10} 
 & WOT & 0.68$\pm$0.03 & 0.38$\pm$0.16 & 0.52$\pm$0.12 & 0.44 $\pm$0.09& \textbf{\uline{0.61$\pm$0.14}} & 0.73$\pm$0.03 & 0.52$\pm$0.12 & \textbf{\uline{0.62$\pm$0.05}} &  & \tabularnewline
\hline 
\multirow{5}{*}{PC1} & DT & 0.90$\pm$0.01 & 0.25$\pm$0.05 & 0.27$\pm$0.05 & 0.26$\pm$0.04 & 0.50$\pm$0.04 & 0.94$\pm$0.03 & 0.27$\pm$0.05 & 0.61$\pm$0.02 & \multirow{5}{*}{Maj: 1032 Min: 77} & \multirow{5}{*}{13.4}\tabularnewline
\cline{2-10} \cline{3-10} \cline{4-10} \cline{5-10} \cline{6-10} \cline{7-10} \cline{8-10} \cline{9-10} \cline{10-10} 
 & S & 0.87$\pm$0.02 & 0.22$\pm$0.04 & 0.38$\pm$0.05 & 0.27$\pm$0.04 & 0.58$\pm$0.03 & 0.90$\pm$0.03 & 0.38$\pm$0.05 & 0.64$\pm$0.02 &  & \tabularnewline
\cline{2-10} \cline{3-10} \cline{4-10} \cline{5-10} \cline{6-10} \cline{7-10} \cline{8-10} \cline{9-10} \cline{10-10} 
 & A & 0.87$\pm$0.02 & 0.26$\pm$0.04 & \textbf{\uline{0.51$\pm$0.06}} & \textbf{\uline{0.35$\pm$0.03}} & \textbf{\uline{0.68$\pm$0.03}} & 0.90$\pm$0.04 & \textbf{\uline{0.51$\pm$0.06}} & \textbf{\uline{0.71$\pm$0.02}} &  & \tabularnewline
\cline{2-10} \cline{3-10} \cline{4-10} \cline{5-10} \cline{6-10} \cline{7-10} \cline{8-10} \cline{9-10} \cline{10-10} 
 & SM & 0.82$\pm$0.05 & 0.16$\pm$0.02 & 0.41$\pm$0.04 & 0.23$\pm$0.03 & 0.59$\pm$0.07 & 0.85$\pm$0.08 & 0.41$\pm$0.04 & 0.63$\pm$0.02 &  & \tabularnewline
\cline{2-10} \cline{3-10} \cline{4-10} \cline{5-10} \cline{6-10} \cline{7-10} \cline{8-10} \cline{9-10} \cline{10-10} 
 & WOT & \textbf{\uline{0.91$\pm$0.03}} & \textbf{\uline{0.34$\pm$0.03}} & 0.30$\pm$0.07 & 0.32$\pm$0.03 & 0.53$\pm$0.02 & \textbf{\uline{0.96$\pm$0.03}} & 0.30$\pm$0.07 & 0.63$\pm$0.02 &  & \tabularnewline
\hline 
\multirow{5}{*}{Heart} & DT & 0.77$\pm$0.03 & \textbf{\uline{0.68$\pm$0.06}} & 0.63$\pm$0.08 & 0.65$\pm$0.05 & 0.73$\pm$0.04 & 0.84$\pm$0.05 & 0.63$\pm$0.08 & 0.74$\pm$0.03 & \multirow{5}{*}{Maj: 188 Min: 106} & \multirow{5}{*}{1.6}\tabularnewline
\cline{2-10} \cline{3-10} \cline{4-10} \cline{5-10} \cline{6-10} \cline{7-10} \cline{8-10} \cline{9-10} \cline{10-10} 
 & S & 0.76$\pm$0.03 & 0.67$\pm$0.06 & 0.57$\pm$0.07 & 0.62$\pm$0.04 & 0.70$\pm$0.04 & \textbf{\uline{0.85$\pm$0.05}} & 0.57$\pm$0.07 & 0.71$\pm$0.03 &  & \tabularnewline
\cline{2-10} \cline{3-10} \cline{4-10} \cline{5-10} \cline{6-10} \cline{7-10} \cline{8-10} \cline{9-10} \cline{10-10} 
 & A & \textbf{\uline{0.79$\pm$0.03}} & \textbf{\uline{0.68$\pm$0.05}} & \textbf{\uline{0.75$\pm$0.06}} & \textbf{\uline{0.71$\pm$0.04}} & \textbf{\uline{0.78$\pm$0.03}} & 0.81$\pm$0.04 & \textbf{\uline{0.75$\pm$0.06}} & \textbf{\uline{0.78$\pm$0.03}} &  & \tabularnewline
\cline{2-10} \cline{3-10} \cline{4-10} \cline{5-10} \cline{6-10} \cline{7-10} \cline{8-10} \cline{9-10} \cline{10-10} 
 & SM & 0.70$\pm$0.03 & 0.55$\pm$0.05 & \textbf{\uline{0.75$\pm$0.06}} & 0.63$\pm$0.03 & 0.71$\pm$0.03 & 0.68$\pm$0.06 & \textbf{\uline{0.75$\pm$0.06}} & 0.71$\pm$0.03 &  & \tabularnewline
\cline{2-10} \cline{3-10} \cline{4-10} \cline{5-10} \cline{6-10} \cline{7-10} \cline{8-10} \cline{9-10} \cline{10-10} 
 & WOT & 0.74$\pm$0.03 & 0.60$\pm$0.06 & 0.74$\pm$0.06 & 0.66$\pm$0.04 & 0.74$\pm$0.03 & 0.75$\pm$0.06 & 0.74$\pm$0.06 & 0.74$\pm$0.03 &  & \tabularnewline
\hline 
\multirow{5}{*}{Segment} & DT & \textbf{\uline{0.96$\pm$0.00}} & \textbf{\uline{0.88$\pm$0.04}} & 0.88$\pm$0.03 & \textbf{\uline{0.88$\pm$0.02}} & \textbf{\uline{0.93$\pm$0.02}} & \textbf{\uline{0.98$\pm$0.00}} & 0.88$\pm$0.03 & \textbf{\uline{0.93$\pm$0.01}} & \multirow{5}{*}{Maj: 1980 Min: 330} & \multirow{5}{*}{6.0}\tabularnewline
\cline{2-10} \cline{3-10} \cline{4-10} \cline{5-10} \cline{6-10} \cline{7-10} \cline{8-10} \cline{9-10} \cline{10-10} 
 & S & \textbf{\uline{0.96$\pm$0.00}} & 0.87$\pm$0.03 & 0.85$\pm$0.03 & 0.86$\pm$0.02 & 0.91$\pm$0.01 & \textbf{\uline{0.98$\pm$0.00}} & 0.85$\pm$0.03 & 0.91$\pm$0.01 &  & \tabularnewline
\cline{2-10} \cline{3-10} \cline{4-10} \cline{5-10} \cline{6-10} \cline{7-10} \cline{8-10} \cline{9-10} \cline{10-10} 
 & A & \textbf{\uline{0.96$\pm$0.00}} & \textbf{\uline{0.88$\pm$0.03}} & 0.87$\pm$0.03 & 0.87$\pm$0.02 & 0.92$\pm$0.01 & \textbf{\uline{0.98$\pm$0.00}} & 0.87$\pm$0.03 & 0.92$\pm$0.01 &  & \tabularnewline
\cline{2-10} \cline{3-10} \cline{4-10} \cline{5-10} \cline{6-10} \cline{7-10} \cline{8-10} \cline{9-10} \cline{10-10} 
 & SM & 0.95$\pm$0.00 & 0.80$\pm$0.03 & 0.87$\pm$0.02 & 0.83$\pm$0.02 & 0.92$\pm$0.01 & 0.96$\pm$0.00 & 0.87$\pm$0.02 & 0.92$\pm$0.01 &  & \tabularnewline
\cline{2-10} \cline{3-10} \cline{4-10} \cline{5-10} \cline{6-10} \cline{7-10} \cline{8-10} \cline{9-10} \cline{10-10} 
 & WOT & 0.72$\pm$0.09 & 0.34$\pm$0.08 & \textbf{\uline{0.97$\pm$0.07}} & 0.51$\pm$0.08 & 0.81$\pm$0.06 & 0.68$\pm$0.11 & \textbf{\uline{0.97$\pm$0.07}} & 0.82$\pm$0.05 &  & \tabularnewline
\hline 
\multirow{5}{*}{Yeast} & DT & 0.83$\pm$0.01 & 0.46$\pm$0.03 & 0.59$\pm$0.05 & 0.51$\pm$0.03 & 0.72$\pm$0.03 & \textbf{\uline{0.87$\pm$0.01}} & 0.59$\pm$0.05 & 0.73$\pm$0.02 & \multirow{5}{*}{Maj: 1240 Min: 244} & \multirow{5}{*}{5.1}\tabularnewline
\cline{2-10} \cline{3-10} \cline{4-10} \cline{5-10} \cline{6-10} \cline{7-10} \cline{8-10} \cline{9-10} \cline{10-10} 
 & S & 0.81$\pm$0.01 & 0.41$\pm$0.04 & 0.60$\pm$0.04 & 0.49$\pm$0.03 & 0.71$\pm$0.02 & 0.85$\pm$0.02 & 0.60$\pm$0.04 & 0.72$\pm$0.02 &  & \tabularnewline
\cline{2-10} \cline{3-10} \cline{4-10} \cline{5-10} \cline{6-10} \cline{7-10} \cline{8-10} \cline{9-10} \cline{10-10} 
 & A & 0.82$\pm$0.01 & 0.43$\pm$0.03 & 0.66$\pm$0.05 & 0.52$\pm$0.02 & 0.75$\pm$0.03 & 0.84$\pm$0.01 & 0.66$\pm$0.05 & 0.75$\pm$0.02 &  & \tabularnewline
\cline{2-10} \cline{3-10} \cline{4-10} \cline{5-10} \cline{6-10} \cline{7-10} \cline{8-10} \cline{9-10} \cline{10-10} 
 & SM & 0.70$\pm$0.02 & 0.32$\pm$0.02 & \textbf{\uline{0.82$\pm$0.03}} & 0.46$\pm$0.02 & 0.75$\pm$0.01 & 0.68$\pm$0.03 & \textbf{\uline{0.82$\pm$0.03}} & 0.75$\pm$0.01 &  & \tabularnewline
\cline{2-10} \cline{3-10} \cline{4-10} \cline{5-10} \cline{6-10} \cline{7-10} \cline{8-10} \cline{9-10} \cline{10-10} 
 & WOT & \textbf{\uline{0.84$\pm$0.02}} & \textbf{\uline{0.50$\pm$0.05}} & 0.73$\pm$0.05 & \textbf{\uline{0.59$\pm$0.03}} & \textbf{\uline{0.79$\pm$0.02}} & \textbf{\uline{0.87$\pm$0.03}} & 0.73$\pm$0.05 & \textbf{\uline{0.80$\pm$0.02}} &  & \tabularnewline
\hline 
\multirow{5}{*}{Oil} & DT & 0.93$\pm$0.01 & 0.35$\pm$0.11 & 0.48$\pm$0.13 & 0.41$\pm$0.11 & 0.68$\pm$0.12 & 0.96$\pm$0.01 & 0.48$\pm$0.13 & 0.72$\pm$0.06 & \multirow{5}{*}{Maj: 896 Min: 41} & \multirow{5}{*}{21.9}\tabularnewline
\cline{2-10} \cline{3-10} \cline{4-10} \cline{5-10} \cline{6-10} \cline{7-10} \cline{8-10} \cline{9-10} \cline{10-10} 
 & S & 0.91$\pm$0.01 & 0.26$\pm$0.07 & 0.48$\pm$0.11 & 0.33$\pm$0.07 & 0.67$\pm$0.08 & 0.93$\pm$0.01 & 0.48$\pm$0.11 & 0.70$\pm$0.05 &  & \tabularnewline
\cline{2-10} \cline{3-10} \cline{4-10} \cline{5-10} \cline{6-10} \cline{7-10} \cline{8-10} \cline{9-10} \cline{10-10} 
 & A & 0.89$\pm$0.01 & 0.22$\pm$0.09 & \textbf{\uline{0.52$\pm$0.11}} & 0.31$\pm$0.08 & 0.69$\pm$0.09 & 0.91$\pm$0.01 & \textbf{\uline{0.52$\pm$0.11}} & 0.71$\pm$0.05 &  & \tabularnewline
\cline{2-10} \cline{3-10} \cline{4-10} \cline{5-10} \cline{6-10} \cline{7-10} \cline{8-10} \cline{9-10} \cline{10-10} 
 & SM & 0.94$\pm$0.01 & 0.41$\pm$0.07 & \textbf{\uline{0.52$\pm$0.10}} & \textbf{\uline{0.46$\pm$0.06}} & \textbf{\uline{0.71$\pm$0.06}} & 0.96$\pm$0.02 & \textbf{\uline{0.52$\pm$0.10}} & \textbf{\uline{0.74$\pm$0.05}} &  & \tabularnewline
\cline{2-10} \cline{3-10} \cline{4-10} \cline{5-10} \cline{6-10} \cline{7-10} \cline{8-10} \cline{9-10} \cline{10-10} 
 & WOT & \textbf{\uline{0.95$\pm$0.02}} & \textbf{\uline{0.47$\pm$0.13}} & 0.28$\pm$0.16 & 0.35$\pm$0.09 & 0.52$\pm$0.12 & \textbf{\uline{0.98$\pm$0.02}} & 0.28$\pm$0.16 & 0.63$\pm$0.07 &  & \tabularnewline
\hline 
\multirow{5}{*}{Adult} & DT & 0.75$\pm$0.00 & 0.48$\pm$0.00 & 0.47$\pm$0.00 & 0.48$\pm$0.00 & 0.63$\pm$0.00 & 0.84$\pm$0.00 & 0.47$\pm$0.00 & 0.66$\pm$0.00 & \multirow{5}{*}{Maj: 37155 Min: 11687} & \multirow{5}{*}{3.2}\tabularnewline
\cline{2-10} \cline{3-10} \cline{4-10} \cline{5-10} \cline{6-10} \cline{7-10} \cline{8-10} \cline{9-10} \cline{10-10} 
 & S & 0.70$\pm$0.00 & 0.41$\pm$0.00 & 0.56$\pm$0.00 & 0.48$\pm$0.00 & 0.65$\pm$0.00 & 0.75$\pm$0.00 & 0.56$\pm$0.00 & 0.65$\pm$0.00 &  & \tabularnewline
\cline{2-10} \cline{3-10} \cline{4-10} \cline{5-10} \cline{6-10} \cline{7-10} \cline{8-10} \cline{9-10} \cline{10-10} 
 & A & 0.71$\pm$0.00 & 0.42$\pm$0.00 & 0.57$\pm$0.00 & 0.48$\pm$0.00 & 0.65$\pm$0.00 & 0.75$\pm$0.00 & 0.57$\pm$0.00 & 0.66$\pm$0.00 &  & \tabularnewline
\cline{2-10} \cline{3-10} \cline{4-10} \cline{5-10} \cline{6-10} \cline{7-10} \cline{8-10} \cline{9-10} \cline{10-10} 
 & SM & \textbf{\uline{0.81$\pm$0.00}} & \textbf{\uline{0.62$\pm$0.01}} & 0.55$\pm$0.01 & \textbf{\uline{0.58$\pm$0.00}} & 0.70$\pm$0.01 & \textbf{\uline{0.90$\pm$0.00}} & 0.55$\pm$0.01 & \textbf{\uline{0.72$\pm$0.00}} &  & \tabularnewline
\cline{2-10} \cline{3-10} \cline{4-10} \cline{5-10} \cline{6-10} \cline{7-10} \cline{8-10} \cline{9-10} \cline{10-10} 
 & WOT & 0.75$\pm$0.02 & 0.48$\pm$0.03 & \textbf{\uline{0.67$\pm$0.05}} & 0.56$\pm$0.01 & \textbf{\uline{0.72$\pm$0.01}} & 0.77$\pm$0.05 & \textbf{\uline{0.67$\pm$0.05}} & \textbf{\uline{0.72$\pm$0.01}} &  & \tabularnewline
\hline 
\multirow{5}{*}{Satimage} & DT & \textbf{\uline{0.91 $\pm$ 0.00}} & \textbf{\uline{0.53$\pm$0.02}} & 0.51$\pm$0.03 & 0.52$\pm$0.02 & 0.70$\pm$0.02 & \textbf{\uline{0.95$\pm$0.00}} & 0.51$\pm$0.03 & 0.73$\pm$0.01 & \multirow{5}{*}{Maj: 5805 Min: 625} & \multirow{5}{*}{9.3}\tabularnewline
\cline{2-10} \cline{3-10} \cline{4-10} \cline{5-10} \cline{6-10} \cline{7-10} \cline{8-10} \cline{9-10} \cline{10-10} 
 & S & 0.90$\pm$0.00 & 0.51$\pm$0.02 & 0.63$\pm$0.02 & 0.56$\pm$0.01 & 0.77$\pm$0.01 & 0.93$\pm$0.00 & 0.63$\pm$0.02 & 0.78$\pm$0.01 &  & \tabularnewline
\cline{2-10} \cline{3-10} \cline{4-10} \cline{5-10} \cline{6-10} \cline{7-10} \cline{8-10} \cline{9-10} \cline{10-10} 
 & A & 0.89$\pm$0.00 & 0.45$\pm$0.03 & 0.60$\pm$0.03 & 0.52$\pm$0.02 & 0.75$\pm$0.01 & 0.92$\pm$0.00 & 0.60$\pm$0.03 & 0.76$\pm$0.01 &  & \tabularnewline
\cline{2-10} \cline{3-10} \cline{4-10} \cline{5-10} \cline{6-10} \cline{7-10} \cline{8-10} \cline{9-10} \cline{10-10} 
 & SM & 0.90$\pm$0.00 & 0.49$\pm$0.02 & 0.72$\pm$0.02 & \textbf{\uline{0.58$\pm$0.01}} & 0.81$\pm$0.00 & 0.92$\pm$0.02 & 0.72$\pm$0.01 & \textbf{\uline{0.82$\pm$0.01}} &  & \tabularnewline
\cline{2-10} \cline{3-10} \cline{4-10} \cline{5-10} \cline{6-10} \cline{7-10} \cline{8-10} \cline{9-10} \cline{10-10} 
 & WOT & 0.88$\pm$0.01 & 0.42$\pm$0.03 & \textbf{\uline{0.75$\pm$0.03}} & 0.54$\pm$0.02 & \textbf{\uline{0.82$\pm$0.01}} & 0.89$\pm$0.01 & \textbf{\uline{0.75$\pm$0.03}} & \textbf{\uline{0.82$\pm$0.01}} &  & \tabularnewline
\hline 
\multirow{5}{*}{Forest cover} & DT & \textbf{\uline{0.97$\pm$0.00}} & \textbf{\uline{0.81$\pm$0.01}} & 0.82$\pm$0.01 & \textbf{\uline{0.82$\pm$0.01}} & 0.90$\pm$0.00 & \textbf{\uline{0.99$\pm$0.00}} & 0.82$\pm$0.01 & 0.90$\pm$0.00 & \multirow{5}{*}{Maj: 35754 Min: 2747} & \multirow{5}{*}{13.0}\tabularnewline
\cline{2-10} \cline{3-10} \cline{4-10} \cline{5-10} \cline{6-10} \cline{7-10} \cline{8-10} \cline{9-10} \cline{10-10} 
 & S & \textbf{\uline{0.97$\pm$0.00}} & 0.78$\pm$0.01 & 0.85$\pm$0.01 & 0.81$\pm$0.00 & 0.91$\pm$0.00 & 0.98$\pm$0.00 & 0.85$\pm$0.01 & \textbf{\uline{0.92 $\pm$0.00}} &  & \tabularnewline
\cline{2-10} \cline{3-10} \cline{4-10} \cline{5-10} \cline{6-10} \cline{7-10} \cline{8-10} \cline{9-10} \cline{10-10} 
 & A & \textbf{\uline{0.97$\pm$0.00}} & 0.79$\pm$0.01 & 0.86$\pm$0.01 & \textbf{\uline{0.82$\pm$0.01}} & \textbf{\uline{0.92$\pm$0.00}} & 0.98$\pm$0.00 & 0.86$\pm$0.01 & \textbf{\uline{0.92$\pm$0.00}} &  & \tabularnewline
\cline{2-10} \cline{3-10} \cline{4-10} \cline{5-10} \cline{6-10} \cline{7-10} \cline{8-10} \cline{9-10} \cline{10-10} 
 & SM & 0.96$\pm$0.00 & 0.73$\pm$0.01 & 0.72$\pm$0.01 & 0.72$\pm$0.00 & 0.84$\pm$0.00 & 0.98$\pm$0.00 & 0.72$\pm$0.01 & 0.85$\pm$0.00 &  & \tabularnewline
\cline{2-10} \cline{3-10} \cline{4-10} \cline{5-10} \cline{6-10} \cline{7-10} \cline{8-10} \cline{9-10} \cline{10-10} 
 & WOT & 0.91$\pm$0.02 & 0.43$\pm$0.05 & \textbf{\uline{0.88$\pm$0.02}} & 0.58$\pm$0.05 & 0.90$\pm$0.01 & 0.91$\pm$0.02 & \textbf{\uline{0.88$\pm$0.02}} & 0.90$\pm$0.01 &  & \tabularnewline
\hline 
\end{tabular}
\label{table:performance}
}
\footnotesize{$^a$ DT=Decision Tree, S=SMOTE, A=ADASYN,  SM=SMOTEBoost, WOT=WOTBoost \\ $^b$ Values are rounded to 2 decimal places}\\
\end{table*}

\begin{figure*}[ht]
  \caption{Performance comparison of G mean and AUC score on 18 datasets}
  \label{fig:comparison}
  \centering
    \includegraphics[width=1.0\textwidth]{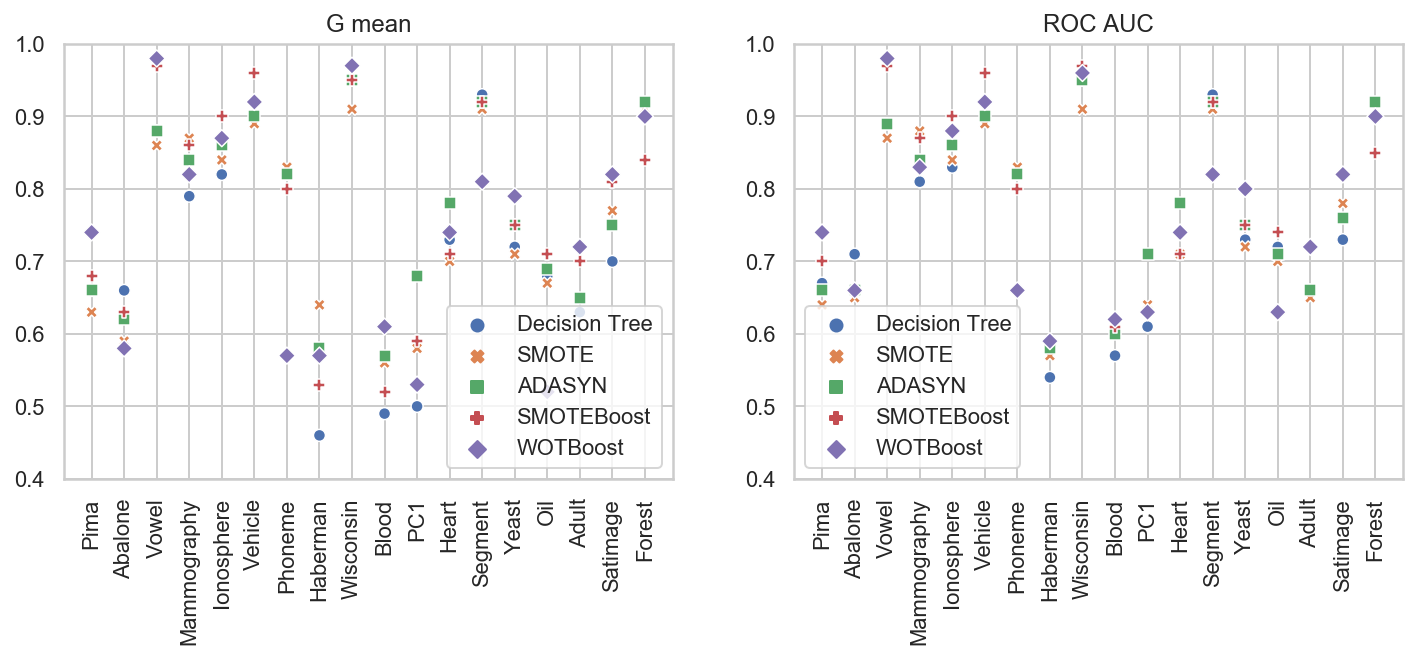}
\end{figure*}

\noindent We highlight the best model and its performance in boldface for each dataset in table \ref{table:performance}. Figure \ref{fig:comparison} presents the performance comparison of these 5 models on G mean and AUC score in 18 datasets. To assess the effectiveness of the proposed algorithm on these imbalanced datasets, we count the cases when WOTBoost algorithm outperforms or matches other models on each metric. The results presented in table \ref{table:winning} show that WOTBoost algorithm has the most winning times on \textit{G mean} (6 times) and \textit{AUC} (7 times). As defined in equation \ref{eq:gmean} in the metric section, G mean is the square root of the product between positive accuracy (i.e., recall or sensitivity) and negative accuracy (i.e., specificity). Meanwhile, area under the ROC curve, or AUC, is typically used for model selection, and it examines the true positive rate and false positive rate at various thresholds.  Hence, both evaluation metrics consider the accuracy of both classes. Therefore, we argue that WOTBoost indeed improves the learning on the minority class while keeping the accuracy of the majority class.\\

\begin{table}[htbp!]
\centering
\caption{Summary of effectiveness of WOTBoost algorithm on 18 datasets}
\resizebox{0.5\textwidth}{!}{
\begin{tabular}{|l|l|l|l|l|l|l|}
 \hline
 Winning counts& Precision & Recall & F\_measure & G\_mean & Specificity & AUC \\ \hline
 Decision Tree & 9 & 0 & 2 & 2 & 11 & 2 \\ \hline
 SMOTE & 2 & 1 & 1 & 3 & 2 & 3 \\ \hline
 ADASYN & 2 & 3 & 3 & 3 & 1 & 3 \\ \hline
 SMOTEBoost & 3 & 10 & 8 & 4 & 2 & 6 \\ \hline
 WOTBoost & 6 & 8 & 4 & 6 & 7 & 7 \\ \hline
\end{tabular}
}
\label{table:winning}
\end{table}

\begin{figure}
  \caption{(a) Distribution of Pima Indian Diabetes dataset. (b) Distribution of Ionosphere dataset}
  \label{fig:distribution}
  \centering
    \includegraphics[width=0.5\textwidth]{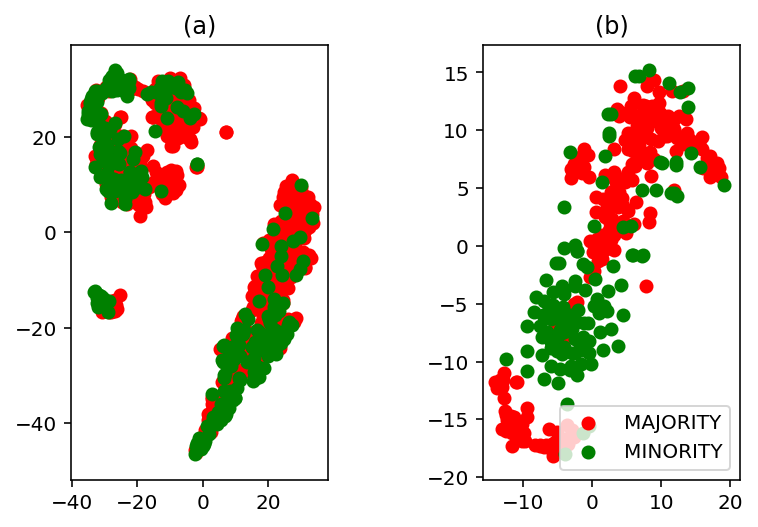}
\end{figure}

\begin{figure}
  \caption{Pima Indian Diabetes distribution before and after applying WOTBoost}
  \label{fig:zoom}
  \centering
    \includegraphics[width=0.5\textwidth]{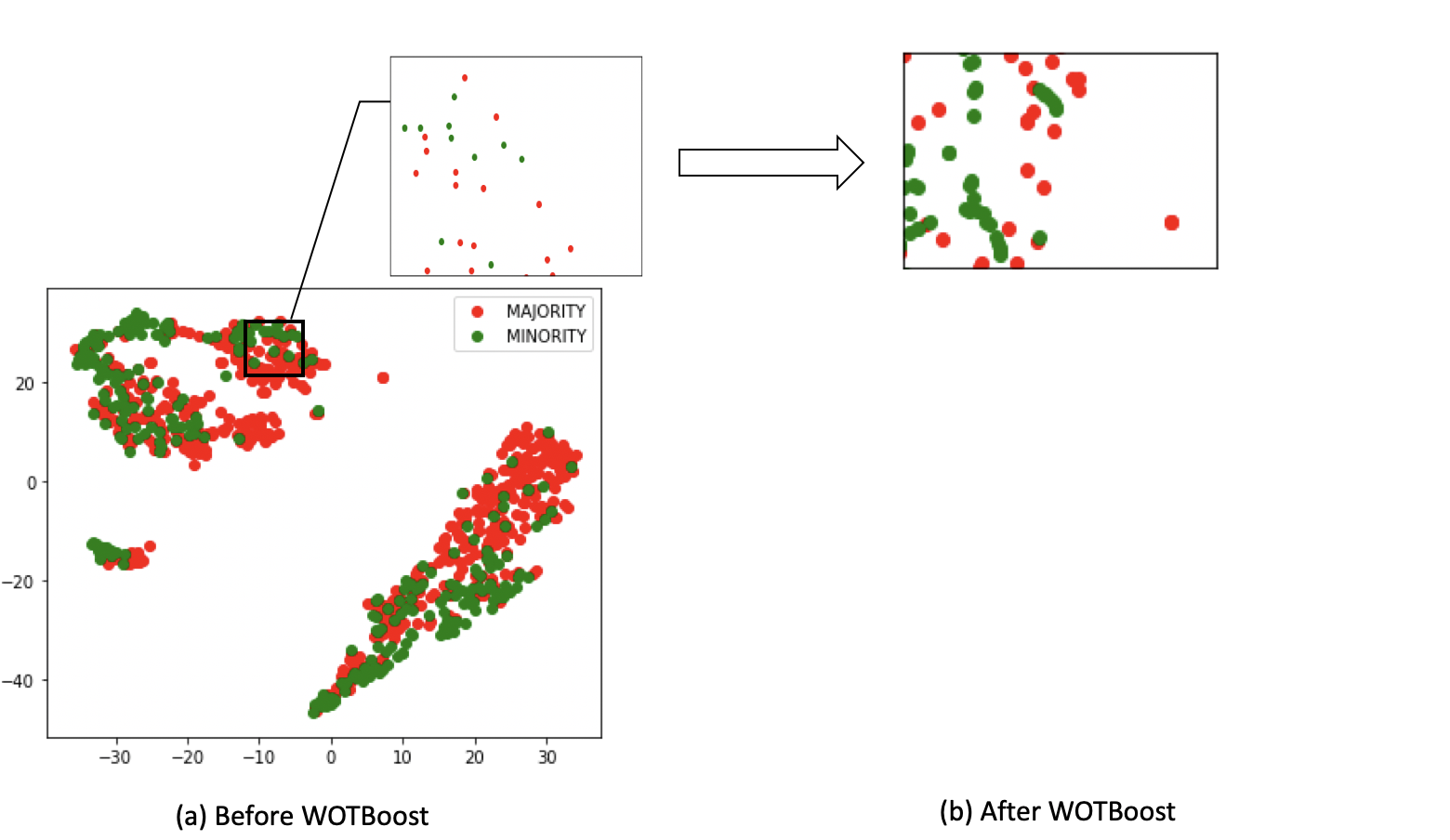}
\end{figure}

In table \ref{table:performance}, we observe that WOTBoost has the best G mean and AUC score on Pima Indian Diabetes whereas SMOTEBoost is the winner on Ionoshphere with the same assessments. Considering these two datasets have similar global imbalanced ratio, it naturally raises the question: are there any other factors that are influential in the classification performance? To understand the reasons why WOTBoost performs better on certain datasets, we investigate the local characteristics of the minority class in these datasets. We use t-SNE to visualize the distribution of these two datasets as shown in Figure \ref{fig:distribution}. t-SNE algorithm allows us to visualize high dimensional datasets by projecting it into a two-dimensional panel. Figure 3 indicates there are more overlapping between two classes in Pima Indian Diabetes, whereas more "safe" minority class samples in Ionosphere. It is likely that WOTBoost is able to learn better when there are more difficult minority data examples. Figure \ref{fig:zoom} demonstrates the distribution of Pima Indian Diabetes before and after applying WOTBoost. We highlight one of the regions where minority data samples are difficult to learn. WOTBoost algorithm is able to populate synthetic data for these minority data samples. \\

\noindent Table \ref{table:data} shows the number of safe/unsafe minority samples of 18 dataset. We consider a minority class sample to be safe if its 5 nearest neighbors contain at most 1 majority class sample; Otherwise, it is labeled as an unsafe minority \cite{han2005borderline,napierala2016types}. Unsafe minority percentage is computed by \[unsafe\ minority\% = \frac{counts\ of\ unsafe\ minority}{counts\ of\ minority}\]\\
\noindent We observe that the unsafe minority percentages are around 50\% or higher in most of the datasets where WOTBoost has the best G-mean or AUC shown in table \ref{table:performance}. For example, Adult, Haberman, Blood Transfusion, Pima Indian Diabetes, and Satimage have 92.5\%, 90.1\%, 87.1\%, 67.9\%, 47.5\% unsafe minority among the total minority class samples, respectively. Meanwhile, the global imbalanced ratios of these datasets are from 1.9 to 10.0. Hence, WOTBoost might be a good candidate to tackle imbalanced datasets with large proportion of unsafe minority samples and relatively high between-class imbalance ratios.\\

\section{Conclusion}

In this paper, we propose the WOTBoost algorithm to better learn from imbalanced datasets. The goal is to improve the performance of classification on minority class without sacrificing the accuracy of the majority class. We carry out a comprehensive comparison between WOTBoost algorithm and 4 other classification models. Results indicate that WOTBoost has the best G mean and AUC scores in 6 out of 18 datasets. WOTBoost shows more balanced performance, such as in G mean, than other classification models compared to particularly SMOTEBoost. Even though WOTBoost is not a cure-all method to the imbalanced learning problem, it is likely to produce promising results for datasets that contain a large portion of unsafe minority samples and maybe relatively high global imbalanced ratios. We hope that our contribution to this research domain would provide more insights and directions.\\

\noindent In addition, our study demonstrates that having the prior knowledge of the minority class distribution could facilitate the learning performance of the classifiers \cite{provost2000machine,guo2004learning,he2008adasyn,napierala2016types,han2005borderline,bunkhumpornpat2009safe}. Further investigating on the data-driven sampling may produce interesting findings in this domain.

\bibliographystyle{unsrt}
\bibliography{references}

\end{document}